# Persistence, Change, and the Integration of Objects and Processes in the Framework of the General Formal Ontology

## Heinrich Herre

2.12.2013

*Abstract.* In this paper we discuss various problems associated to temporal phenomena. These problems include persistence and change, the integration of objects and processes, and truth-makers for temporal propositions. We propose an approach which interprets persistence as a phenomenon emanating from the activity of the mind, and which, additionally, postulates that persistence, finally, rests on personal identity.

The General Formal Ontology (GFO) is a top level ontology being developed at the University of Leipzig. Top level ontologies can be roughly divided into 3D-Ontologies, and 4D-ontologies. GFO is the only top level ontology, used in applications, which is a 4D-ontology admitting additionally 3D objects. Objects and processes are integrated in a natural way.

## 1. Introduction

This paper is devoted to the topic of identity over time and to problems arising from this phenomenon. If a thing really changes then it cannot be identical before and after the change. From this condition follows by contraposition, that no thing has undergone any change. On the other hand, we usually assume the existence of changing things, being identical over time. There are various approaches to solve this puzzle. Aristotle, for example, distinguished between accidental and essential properties of a thing. Then, the change of an accidental property preserves the thing's identity, whereas the change of an essential property leads to a new thing. This solution of the puzzle depends on a clear distinction between accidental and essential properties. This distinction is controversial and various solutions were developed that do not require this distinction. But it turns out that any of these solutions exhibit some weakness.

A solution of this problem is approached within GFO by a deeper analysis of the notion of "thing". GFO advocates the thesis that the phenomenon of identity of changing things arises from the integration of three different pair-wise disjoint kinds of entities. These entities are called in GFO presentials, continuants, and processes. Processes cannot change as a whole, though, they can possess changes; a change of a process is a kind of discontinuity of this process. Continuants persists through time, preserving their identity, and may change through time. Since continuants remain identical over time, the changes of them are realized by another kind of entities, called presentials. Continuants are creations of the mind, and possess, according to GFO, not the same level of objectivity as processes and presentials. We defend the thesis that these subjective creations rest, finally, on the phenomenon of personal identity and the continuity of time; hence, the open problems of identity over time for arbitrary entities, are finally reduced to the problem of personal identity. The present paper continues the research of [Ba12] and is focused on the problem of persistence and the temporal incidence problem.

In section 2 we survey problems and basic approaches to time and temporal phenomena. Section 3 devoted to the basics of GFO, and section 4 presents an ontology of functions, whereas section 5 gives an overview of properties of processes. In section 6 the principles of integrative realism are expounded, as adopted and invented by GFO as a basic doctrine. In section 7 the basic integration axiom of GFO is presented and discussed; it is defended that identity over time rests, finally, on personal identity, from which the creation of persisting, changing entities, called continuants, emanates. In section 8 the conclusion and some further problems are discussed.

## 2. Time and Temporal Phenomena

Humans access time through phenomena of duration, persistence, happening, non-simultaneity, order, past, present and future, and change. From this manifold of temporal phenomena pure time, called phenomenal time, is abstracted [Ba12]. We hold that this phenomenal time and its associated temporal phenomena are mind-dependent. On the other hand, we assume that material entities possess mind-independent dispositions to generate these temporal phenomena. We call these dispositions temporality and claim that they unfold in the mind as a manifold of temporal phenomena. This distinction between the temporality of material entities and the temporal phenomena corresponds to the distinction between temporality and time as considered by Nicolai Hartmann in [Ha65]. A satisfactory ontology of time and its formal representation should treat the various temporal phenomena in a uniform und consistent manner. In [Ba12] four basic topics are suggested, and a complete axiomatisation of phenomenal time, in the spirit of Brentano's approach, was presented. In the present paper, we focus more precisely on the following three topics.

(1) Persistence and change
(2) Integration of objects and processes.
(3) Truth-makers for propositions

*The topic (1)* lacks a comprehensive and generally accepted theory. There are several approaches to cope with this phenomenon. One approach relates to an endurance/perdurance distinction. This terminology was introduced by Mark Johnston, and described by David Lewis in the "Plurality of Worlds."[Le86]. David Lewis classifies entities into endurants and perdurants [Le86]. An entity perdures if it persists by having different temporal parts, or stages, at different times, whereas an entity endures if it persists by being wholly present at any time of its existence. Persistence by endurance is paradoxical and leads to inconsistencies [BD05]. The stage-approach exhibits serious weaknesses [Wa08]. Both approaches are criticized by various arguments, while further alternatives, e.g. in [Hs03], reveal shortcomings, as well. Consequently, the development of a satisfactory, widely acceptable theory of persistence remains an open problem. GFO proposes an approach and claims that the persistence of spatio-temporal individuals is a creation of the mind and is as mysterious as personal identity.

Concerning the *topic (2)* one must distinguish between 3D and 4D ontologies. In general, an ontology, admitting objects and processes, faces the problem how these entities are related. A 3D ontology admits objects as the fundamental category of spatio-temporal entities. Objects are considered as persisting entities that may change. Processes are introduced as a separate category, and it arises the problem how processes are related to objects. Usually, a process in 3D ontologies is understood as an entity that depends on an object, and the relation, connecting both is a kind of inherence relation. This approach has serious drawbacks, in particular in the field of physics, see [Ro13], [Lo10]. In 4D ontologies processes are the fundamental category of spatio-temporal entities, and there are no objects in the sense of 3D ontologies. Objects in a 4D ontology are particular processes, hence, there is no integration problem. On the other hand, pure 4D ontologies do not adequately treat ordinary objects that are actually not processes; hence, they are conceptually incomplete. GFO offers another solution of the problem (2). Since GFO is basically a 4D ontology, objects must be introduced and explained. Objects in GFO are not processes, though, they are naturally related to processes; and this relation is expressed by the integration axiom.

The *topic (3)* concerns the problem which kind of entities make propositions true, and how this truth-relation can be made explicit. We restrict this problem to truth-makers, being spatio-temporal material entities. Then, for propositions φ we must specify a relation |= and introduce entities TM, such that for e in TM: e |= φ. Since in GFO processes are the most fundamental category, we may assume that e grounded on a process. A process exhibits several levels of description, which can be understood as properties of processes. Hence, the condition e |= φ holds, if e is a property of a process, hence e serves a truth-maker for φ. Since φ is closely tied to a process, φ itself can be considered as property of e.

## 3. Basics of GFO

In GFO the entities of the world, being different from sets, are classified into categories and individuals. Categories can be instantiated, individuals are not instantiable. GFO allows for categories of higher order, i.e., there are categories whose instances are themselves categories, for example the category "species". Spatio-temporal individuals are classified along two axes, the first one explicates the individuals' relation to time and space, and the second one describes the individuals' degree of existential independence.

### 3.1 Classification of Individuals based on their relation to space and time

Spatio-temporal individuals are classified into continuants, presentials and processes. *Continuants* persist through time and have a lifetime; they correspond to ordinary objects, as cars, balls, trees etc. The lifetime of a continuant is presented by a time interval of non-zero duration; such time intervals are called chronoids in GFO, see [Ba12]. Continuants are individuals which may change, for example, an individual cat C crossing the street. Then, at every time point t of crossing C exhibits a snapshot C(t); these snapshots differ with respect to their properties. Further, the cat C may loose parts during crossing, though, remaining the same entity. The entities C(t) are individuals of their own, called *presentials*; they are wholly present at a particular time-point, being a time-boundary. Presentials cannot change, because any change needs an extended time interval or two coinciding time-boundaries.

*Processes* are temporally extended entities that happen in time, for example a run; they can never be wholly present at a time-point. Processes have temporal parts, being themselves processes. If a process P is temporally restricted to a time-point then it yields a presential M, which is called a process boundary of P. Hence, presentials have a two different origins, they may be snapshots of continuants or process boundaries. There is a duality between processes and presentials, the latter are wholly present at a time-point whereas this is never true for processes. The corresponding classes/sets of individuals, denoted by the predicates $Cont(x)$, $Pres(x)$, and $Proc(x)$, are assumed to be pair-wise disjoint. Processes present the most important kind of entity, whereas presentials and continuants are derived from them. There are several basic relations which canonically connect processes, presentials, and continuants, see section 7.

### 3.2 Classification of individuals related to the degree of independence and complexity.

Spatio-temporal individuals, according to the second axis, are classified with respect to their complexity and their degree of existential independency. Attributives depend on bearers which can be objects (continuants, presentials) and processes, whereas processes are always parts of situations.

Situations are parts of reality which can be comprehended as a coherent whole [BP 1983]. An example of a situation is a football match with temporal extension and spatial location, which includes various entities such that a coherent whole is established. Situations are classified into temporally extended situations, sometimes called situoids, and presentic situations, being present at time-point. A presentic situation is a snapshot of a football match. Situations are considered as individuals, the specification of which needs universals (in particular relational universals), associated to them. We assume that contexts can be regarded as situations.

There is a variety of types of attributives, among them, qualities, roles, functions, dispositions, and structural features. Categories the instances of which are attributives are called properties throughout this paper. According to the different types of attributives (relational roles, qualities, structural features, individual functions, dispositions, factual, etc.) we distinguish quality properties (or intrinsic properties) and role properties (extrinsic properties), and the role properties are classified into relational role properties (abr. relational properties), social role properties (social properties)(Lo ). etc. The dependency relations between the entities of different types are displayed in the figure 1.

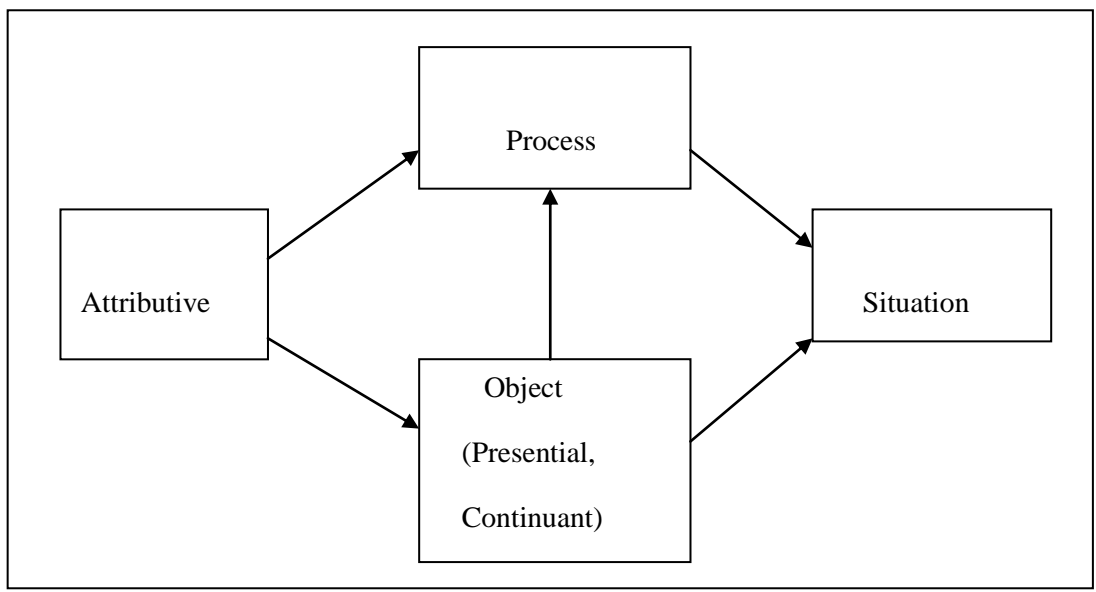

## 4. Functions

An analysis of the different types of attributives is work in progress. Any of these types establishes an ontology of its own. In this section we summarize the ontology of functions of GFO. Functions exhibit important properties of entities which cannot be extracted from measurements, they are mental constructions. A function, in GFO, possesses a psychological entity as a component. This entity is an idea or a thought which is directed to a goal in the future. Psychological entities belong to the psychological stratum [Po01], being one of the four fundamental ontological regions adopted by GFO.

This goal of a function can be understood as a set of situations to be achieved, and this set is described by a category. These goals are imagined in the future (they are, in a sense, anticipated entities); hence, to achieve them, we must start with the existing present situations. These situations, we are starting from, are called requirements. On a more abstract level, a function in OF relates initial situations, called requirements, to situations to be achieved, called goals. A functional item is a set of necessary conditions that an entity must satisfy to be able to achieve the goal within a realization.

The term *function* in OF exhibits various meanings which are made explicit as follows. A function f can be an intentional entity, called *intentional function*, specified by the predicate IntF(f); a function f can be understood as a conceptual structure, called *conceptual function*, and specified by ConcF(f); a function f can be an *individual function*, specified by IndF(f); finally, a function f can be understood as a *universal function*, specified by the predicate UnivF(f). There are relations between these different interpretations of a function which can be expressed by logical formulae.

Throughout the remainder of the paper we use the term *function* in the sense of a *conceptual structure*, the term *universal function* as a concept, and the term *individual function* denotes an individual. In the framework of GFO an individual functions is an attributive, a universal function is a concept, whereas a conceptual structure is a system composed of concepts and sets.

**Definition**. A function *f* is a conceptual structure CStr(f) of the following form:
CStr(*f*)=(Label(*f*), Req(*f*), Goal(*f*), FItem(*f*)), where:
- Label(*f*) denotes a set of *labels* of function *f*, which are natural language expressions, informally describing "to do something".
- Req(*f*) denotes a concept, called *requirements,* the instances of which are parts of material reality, which must be present if the function *f* is to be realized.
- Goal(*f*) denotes a concept the instances of which are parts of material reality being intended by some agent as a result of successful realizations of the function *f*.
- FItem(*f*), called functional item of *f,* is a system of necessary properties which a bearer of *f* must satisfy to execute a realization of *f*.

A function specifies *what* is to be done, whereas the realization of functions exhibit another aspect - it refers to a specification of *how* a goal of the function comes to reality. A realization of a function is an individual spatio-temporal entity which is based on a binary relation, linking a function with an entity to be brought about. Intuitively, we call an *actual realization* of the function *f* an entity which results in an achievement of an individual goal of *f* in the circumstances satisfying an individual requirement of *f*, hence, an actual realization connects an individual requirement of a function *f* with an individual goal of it.

The binary relation $Rl_{act}(x,y)$ has the meaning, *x* is an actual realization for the function *y*. A category *x* is called an actual universal realization of a function *y,* denoted by $UnRl_{act}(x,y)$, if every instance of *x* is a realization of *y*, and furthermore, the instances of *x* cover the requirements of the function.

Often a realization is a complex entity which has a number of entities participating in it. In certain situations some of these entities can be identified as those which execute the function realization. The execution of entity *y* by entity *x*, denoted by the binary relation Exe(*x,y*), is understood as a causal influence that *x* has on *y*[1]. For instance, to the process *p* of blood movement, which is a realization of function *f*: *to pump blood*, contribute the heart *h*, the blood, and the veins. However, the role that the heart *h* has in the process *p* is different from the roles of veins and blood, namely, it is the heart which actually pumps blood and, thus, it can be said that the heart *h* executes the realization *p* of the function *f*. The inter-relations between the mentioned entities are expressed by the fact Exe(*h,p*), saying that the heart *h* executes the process of blood movement *p*. An individual *x*, executing a realization of a function *f*, is called an *actual realizer* of function *f* and is expressed by a relation $R_{Act}(x,f)$.

---

[1] Both the notions of achievement and execution are strongly related to the notion of causality and itself are important and are non trivial problems which however are out of the scope of the current paper and thus taken here as primitive.

## 5. Properties and Truth-Makers

Properties, according to section 3, are categories (concepts), the instances of which are attributives. We defend a minimalist approach to properties and restrict our investigation to properties of material entities, notably, material continuants, presentials, and material processes. A truth-maker for a proposition φ is a part P of reality that satisfies φ (satisfying φ, denoted by P | = φ, called "P makes φ true". The explication of the relation sat(X,Y) must clarify what kind entities X,Y are, and how the relation sat(X,Y) can be clearly described. We advocate that X has the form (P,S,f), where P is a process, and S is a situation, associated to P, and f is a part of S. Since propositions can have a very complex nature, we restrict our investigation to the most simple propositions, called, throughout the paper elementary proposition. Elementary proposition correspond to single facts.

### 5.1 Properties of Processes

Since objects depend on processes, a classification of processual properties include the properties of objects as a special case. We classify properties of processes with respect to two dimensions, the level of abstraction, and the temporal structure. The levels of abstraction are grounded on various types of mental procedures, called intentional acts. The most basic intentional act is a direct perception of a visual situation: A part of the world is comprehended as coherent whole. An analytic intentional act of the mind distinguishes objects, properties, and facts within the situation. Finally, by a meta-reflective intentional act propositions a formed, the content of which related to the entities created by the analytic intentional act.

Let us consider an example. A person sees a scene as a whole, say, a situation of a foot-ball match. The analytic intentional act of this person results in distinguishing individual players, the ball and the goal. Furthermore, the person sees a player shooting the ball to the goal. From this he may create the following fact F: "The player's shooting the ball to the goal", and by a meta-reflective intentional act he may create the following proposition P: "The player is shooting the ball to the goal". The fact F, as a part of reality, is neither true nor false, whereas the proposition P has a truth-value, and the fact F makes the proposition P true. Furthermore, the fact F is not the same as the underlying process; any fact is grounded on a process, and, in a sense, we may say that this fact is a attributive (a property) of the underlying process.

Phenomenal properties can be directly perceived or measured; they exhibit a sublevel of the analytic level. Examples are the size, weight or velocity of an object, or its color or form. Facts are constructed by using relations and relators, which are creations of the mind; the same holds for functions. They cannot be directly perceived or measured, for example, the function of hammering, being ascribed to stone. Finally, the propositional level is created by meta-reflective intentional acts, yielding propositions and background knowledge. We may call this level the knowledge level. Obviously, the knowledge level adds information to the processual information, which cannot be directly extracted from the process. This kind of knowledge can be understood as an interpretation of the process. Any of these levels exhibit a two-sided dependence, on the one hand they depend on the real process, being independent from the subject, on the other hand, it depends on the mind. There is an active interrelation between the objective process and the mind, and this interrelation is explicated on the basis of the integrated realism and ontological pluralism, being a basic assumption of GFO.

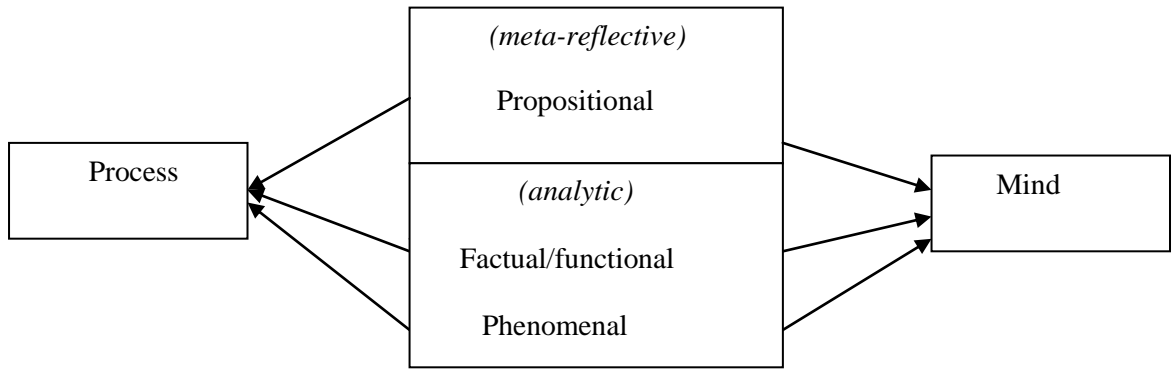

Another dimension for the classification of process properties is the temporal structure associated with the considered property. A property of a process P is presentic if it holds at a process boundary of P. The presentic phenomenal properties are distinguished into isolated and non-isolated properties. For example, the color red of a moving ball B, participating in the process P and considered at a time point, is an isolated property of P, whereas the velocity of B at a time-point is non-isolated, because the determination of the velocity needs a temporally extended interval. A phenomenal property of a process P is global if it cannot be presented at a time point, for example, an electro-cardiogram, viewed as a property of the process of heart activity. Similar distinctions can be made for factual/functional properties and for propositional properties of a process. If a process P is the realisation of a function F, then we say that F is a functional property of P; and this property is global, and hence, non-presentic. The functional structure of a process can be rather complex, because a process can have temporal parts, realising various different functions.

### 5.2 Truth-Makers

The basic idea is that a propositions $\varphi$ is satisfied by an entity (P, S, f), where P is a process, being the foundation of a situation S, and f is a part of S. In the simplest case f is a single fact, hence, $(P,S,f) \models \varphi$. S can be considered as a complex property of P, created by the mind. Typically, f is a property of a process, for example a fact. Obviously, the whole situation S (if temporally extended) is based on the process. We explain this by an example: "John's drinking of a beer" is a part (denotes a part) of reality associated to a process, whereas "John is drinking a beer" is proposition which can be true or false (i.e. has a truth-value).

The same holds for facts. Propositional properties can be classified similarly. A process P has the propositional property $\varphi$ if P satisfies $\varphi$ in the sense that P is a truth-maker for $\varphi$. A proposition refers to the level of factual/functional properties of a process, hence, the relation of a propositional property to the phenomenal level is indirect. Though, usually it is possible to acquire data of the phenomenal level (say measurements) associated to a factual, or function or propositional property.

## 6. Integrative Realism of GFO and Ontological Pluralism

GFO advocates four ontological regions, the material region (including the domains of natural sciences as physics, chemistry, and biology), the psychological region, the socio-systemic region, and the ontological region of ideal entities. This approach draws, among others, on ideas of Hartmann [Ha65] and Poli [Po01]. The region of ideal entities includes Platonic ideas, in particular the mathematical entities, but additionally creations of the mind, among them idealizations in science, model-constructions, and theories. These four regions are interrelated; the psychological and the socio-systemic region is founded on the material region. The mind is an entity which has relations to any of these regions; it might be considered as the basis for the region's integration and unification. The psychological and the socio-systemic region are mutually interrelated and active in both directions. There is a relation between mathematical entities and the physical world which is not yet understood, as emphasized by Wigner [Wi60].

Integrative realism is a theory which is concerned with the interrelation between the mind and the ontological regions of the world, and the role of the mind for establishing an ontology in general. The mind can access the various ontological regions only by subjective constructions, notably by concepts. Integrative realism states conditions about the relation between the mind and the other ontological regions, in particular, the material region. Several phenomena are related to the material region, including matter, occupation of space, change, and movement. Integrative realism postulates the existence of objective dispositions, being inherent in the independent material things, and which need necessarily a mind to be unfolded and to come to reality. The color red, for example, is a subjective phenomenon which is associated to a physical entity e. This entity e has an objective disposition which is unfolded in a mind to become the subjective phenomenon of red. We hold that subjective phenomena cannot be reduced to physical objects. Hence, the unfolding relation bridges two completely different ontological regions, and the subjective region is governed by mind-internal laws which cannot be described by physical laws. A similar approach is expounded in [Ma02], [Ma13].

## 7. Integration of Objects and Processes

### 7.1 The Axiom of Object-Process-Integration

In this section we consider the problem how presentials, continuants, and processes are related, and why we need an integration of these entities. Should ordinary objects be considered as perduring, enduring, or as stages of processes? We hold that we need all three of these entities to achieve a complete picture and understanding of the world. Let us consider the following sentence: "Today, on May, 2013 at 5 pm, John meets Paul with whom he was thirty years ago on January 1983, at time 8 pm, involved in accident." The terms "Paul" and "John" denote spatio-temporal individuals, and if they are related to the mentioned time-points we consider them as presentials. John and Paul are the same persons on May 2013 and February 1983, hence, there must be continuants, denoted by the terms "John" and "Paul", which exhibit the presentic John and Paul at the respective time-points. Presentic entities have no temporal extension, hence, a presentic John or Paul cannot do anything, since any action needs a temporal interval. Furthermore, the presentic Johns and Pauls, exhibited by the corresponding continuants, must be causally connected such they cannot be "replaced" by other presentials. We postulate that there are processes, denoted by "John" and "Paul", which are assumed for action in time and for establishing causal connectedness between the corresponding presentic individuals. From this follows that the terms "John" and "Paul" denote different types of individuals which are needed to achieve a complete interpretation and understanding of these terms.

Processes are the most fundamental class of spatio-temporal individuals, and John, understood as a process P, contains the presentic Johns by restricting the process P to time points of the process' temporal extension. The restriction of a process P to a time-point t is called the process boundary of P at t. According to GFO, a process is not the mereological sum of its boundaries. We postulate that every material presential is a part of a process boundary, and, hence, that presentials have no independent existence, they always depend on a process.

Continuants present the phenomenon of identity over time, and it arises the question where the continuants come from? The problem of identity over time was treated by philosophers already in ancient time, and especially in the European philosophy of the $17^{th}$ and $18^{th}$ century. One idea is to think of persistence of bodies, particularly in motion, as a succession of new entities, or re-creations at different places. Leibniz puts this idea in a letter to Princess Sophie: "The duration of things or the multitude of momentary states is the collection of an infinity of strokes of God, of which each one at each instant of time is a creation or reproduction of everything, without a continuous passage, narrowly speaking, from one state to the next." [2]. According to GFO, continuants are creations of the mind, which are constructed out of presentials. This approach seems to be compatible with Leibniz's idea; the strokes correspond to the presentials, and the continuous passage between the momentary states is an illusion, based on the introspectively accessed continuous flow of time. Though, in contrast to Leibniz's approach a continuant in GFO is additionally underpinned by a process.

Any process has a temporal extension, being a connected time-interval, called chronoid [Ba12]. Given a process P and a temporal part of its temporal extension, then the restriction of P to this subinterval, denoted by P', is called temporal part of P. The restriction of P to time-point of the temporal extension is called process boundary at that time-point. Obviously, every process boundary is a presential, i.e. an entity being wholly present at this time-point. According to GFO, these continuants are cognitive creations, built up from presentials ("snapshots") by our cognitive apparatus in a similar way as seeing a movie in which continuants are created (out of the movie's snapshots).

The integration axiom of GFO states that for every continuant C there exists a process P, the boundaries of which coincide with the presentials, exhibited by C ([He06], [He10]). A formalization of this axiom needs a number of relations: *exhib*(C,t,M) (the continuant exhibits the presential M at time point t), and *procbd*(P,t,N) with the meaning that N is the process boundary of the process P at the time-point t.

**Axiom of Object-Process Integration.**
For every material continuant C there exists a process Proc(C) such that the process boundaries of Proc(C) coincide with the presentials, exhibited by C, formally,

$\forall C \ (MatCont(C) \rightarrow \exists P \ (Proc(P) \land lft(C) = tempext(P) \land \forall t \ M \ (exhib(C,t,M) \leftrightarrow procbd(C,t,M))$.

In comparison to other top level ontologies (such as BFO [Sp06], DOLCE [MBG03], UFO [GW10]), GFO is the only ontology, used in practical applications, for which the processes are the most fundamental category of spatio-temporal individuals, whereas objects and their snapshots (presentials) depend on processes. This integration principle allows for powerful applications; it distinguishes GFO from other process ontologies. The approach of perdurantism by David Lewis [Le85] does not allow for continuants, i.e. concrete individuals having a life time and persisting through time by enduring, and exhibiting at any time point an entity, being wholly present at that time point. Another process ontology is presented by T. Sider in [Si03]. The stages in Sider's approach corresponding to the process boundaries in GFO, though, process stages are different from process boundaries. In the stage theory a process is the mereological sum of its stages, which is not true in GFO. A process cannot be identified with the set (or mereological sum) of its process boundaries.
.

---

[2] This citation is from [Bl99], entry endurance/perdurance.

## 7.2 Continuants and Personal Identity

In this section we address the question of which mental conditions and features enable the mind to create continuants. We postulate that the creation of continuants is based on personal identity and the introspectively accessed continuity of time. Personal identity states that an individual mind, a self, exhibited by consciousness, is identical at different times. But what "different times" means? A time may have a duration, hence it may represent a time interval. We assume that these "times" are time-points, more precise, time-boundaries of the temporal extension of a process. Consciousness as a state of the mind can be located at time points, though, these instantaneous states are boundaries of a continuous stream of consciousness.[3]

An individual mind, a person, feels to be identical at different time-points, and this feature is adapted to a series of perceived snapshots, being presentials. As a result the mind creates a continuant, being an individual which is perceived as identical over time. We hold that the construction of continuants in the case of a movie has the same nature.  There are various approaches to explain personal identity. One attempt is the following memory argument for personal identity which turns out to be circular. "A being the same person as B" in terms of A remembering what happened to B. Though, any plausible analysis of "A remembering what happened to B" will mention, in this analysis, A being the same person as B.  There another approach that the enduring self is a fiction, or a figment of the imagination. We have no reason to think "I" in terms of a single unified self that owns a variety of experiences or states; we have only access to the succession of the states themselves. The enduring self is then a fiction, or a figment of the imagination. But where this imagination comes from? Does this imagination assume already personal identity? The nature of conscious experience has been the largest obstacle to physicalism, behaviourism, and functionalism in the philosophy of mind.

We conclude his section with somewhat speculative remarks. The phenomenon of personal identity may emanate from a sub-conscious level of the self. In Buddhism, there is the notion of atman: the self or soul, conceived of as lying behind the empirical self, and in Hindu thought an eternal unity behind the self is postulated, identified with Brahman. Another understanding of personal identity can be related to the philosophy of A. Schopenhauer. Personal identity is grounded on the sub-conscious will. There are individual wills, which are reflected and mirrored in the conscious component of the self and constitute the person's core. Though, these individual wills are related so some eternal unifying will of the world. In the spirit of Schopenhauer's philosophy we may postulate that the integration of the individual will, the conscious self, and the eternal universal will creates personal identity. The crucial citation is from A. Schopenhauer's *Über den Satz vom Grunde, § 42. Subjekt des Wollens*.

"Die Identität des Subjekts des Wollens mit dem erkennenden Subjekt, vermöge welcher das Wort 'Ich' beide einschließt und bezeichnet*, ist der Weltkknoten und daher unerklärlich*. ... Eine wirkliche Identität des Erkennenden mit dem als wollend Erkannten, also des Subjekts mit dem Objekt, ist unmittelbar gegeben. Wer aber das Unerklärliche dieser Identität sich recht vergegenwärtigt, wird sie mit mir *das Wunder schlechthin* nennen."

---

[3] An literary example of the phenomenon of stream-of-consciousness is mirrored and expressed by Molly Bloom's saying over at the end/ the conclusion of J. Joyce's novel "Ulysses".

## 8. Conclusion and Further Research

In the present paper we expounded and discussed an approach to persistence through time and to the temporal incidence problem. Changing entities, being identical over time, are exemplified by continuants and we state the hypothesis that such entities are creations of the mind. We hold that these creations are based on two phenomena, the personal identity, and the continuity of time, accessed by the subject's introspection. Hence, an ultimate understanding of identity over time is reduced to the problem of personal identity, and to the temporal structure of phenomenal time.

The framework, presented in this paper, can be further developed in several directions. First of all, there is a need to further elaborate a top level ontology of processes. This includes an ontology of process-properties. Furthermore, the integration-axiom exhibits a stable basis for practical applications; it was already applied to model stem cell processes. Concerning personal identity, there is extensive research on this topic in philosophy of mind. We believe that ontological pluralism and the integrative realism may be useful in the investigation of this question.

**Acknowledgement** I thank Frank Loebe for reading the paper and for his corrections and remarks which contributed to the quality of the paper.